\documentclass{article}
\usepackage{cite}
\usepackage{amsmath,amssymb,amsfonts}
\usepackage{algorithmic}
\usepackage{graphicx}
\usepackage{textcomp}
\usepackage[table]{xcolor}
\usepackage{subfig}
\usepackage{hyperref}
\usepackage[ruled,vlined]{algorithm2e}
\usepackage[symbol]{footmisc}
\usepackage{multirow}

\hypersetup{
    colorlinks=true,
    linkcolor=red,
    filecolor=magenta,      
    urlcolor=cyan,
}

\usepackage{geometry}
\usepackage{titlesec}

\makeatletter
\def\@seccntformat#1{\@ifundefined{#1@cntformat}%
  {\csname the#1\endcsname\quad}
 {\csname #1@cntformat\endcsname}}
\makeatother

\renewcommand{\thesection}{\arabic{section}.}
\renewcommand{\thesubsection}{\thesection\arabic{subsection}.}
\renewcommand{\thesubsubsection}{\thesubsection\arabic{subsubsection}.}

\titleformat*{\section}{\Large\bfseries}
\titleformat*{\subsection}{\large\bfseries}
\titleformat*{\subsubsection}{\large\bfseries}
\titleformat*{\paragraph}{\large\bfseries}

\usepackage{times}

\usepackage[table]{xcolor}
\definecolor{lightgray}{rgb}{0.95, 0.95, 0.95}
\definecolor{darkgray}{rgb}{0.85, 0.85, 0.85}

\usepackage[nottoc]{tocbibind}
\usepackage{pifont}
\usepackage{xcolor}
\usepackage{url}
\usepackage{tikz}
\usepackage{authblk}
\usepackage{adjustbox}

\begin{document}
\twocolumn

\title{ \vspace{-12mm} \line(1,0){490} \\
\vspace{2mm}
DeepHAZMAT: Hazardous Materials Sign Detection and Segmentation with Restricted Computational Resources
\\ \vspace{-2mm} \line(1,0){490}  \vspace{-6.7mm} \\\line(1,0){490}
}
\date{}

\author{Amir Sharifi $^{1, \, \star}$, \, Ahmadreza Zibaei $^{2, \, \star}$, \,  Mahdi Rezaei $^{3, \, \dagger}$\\
$^{1, 2}$ Advanced Mobile Robotics Lab, Qazvin Azad University, Qazvin, IR\\
$^1$ \href{mailto:amir.sharifi@qiau.ac.ir}{amir.sharifi@qiau.ac.ir}  \hspace{5mm} $^2 $\href{mailto:a.zibaei@qiau.ac.ir}{a.zibaei@qiau.ac.ir}\\
$^3$ Faculty of Environment, Institute for Transport Studies, The University of Leeds, Leeds, UK\\
$^3$ \href{mailto:m.rezaei@leeds.ac.uk}{m.rezaei@leeds.ac.uk}}

\makeatother
\maketitle

\begin{abstract}
\vspace{-1mm}
One of the most challenging and non-trivial tasks in robot-based rescue operations is the Hazardous Materials or HAZMATs sign detection in the operation field, to prevent further unexpected disasters. Each Hazmat sign has a specific meaning that the rescue robot should detect and interpret it to take a safe action, accordingly. Accurate Hazmat detection and real-time processing are the two most important factors in such robotics applications. Furthermore, we also have to cope with some secondary challenges such as image distortion and restricted CPU and computational resources which are embedded in a rescue robot. In this paper, we propose a CNN-Based pipeline called DeepHAZMAT for detecting and segmenting Hazmats in four steps; 1) optimising the number of input images that are fed into the CNN network, 2) using the YOLOv3-tiny structure to collect the required visual information from the hazardous areas, 3) Hazmat sign segmentation and separation from the background using GrabCut technique, and 4) post-processing the result with morphological operators and convex hull algorithm. In spite of the utilisation of a very limited memory and CPU resources, the experimental results show the proposed method has successfully maintained a better performance in terms of detection-speed and detection-accuracy, compared with the state-of-the-art methods.\\
\end{abstract}

\vspace{-2mm}
\textbf{Keywords:} Hazardous Materials, Hazmat Sign Detection, HAZMAT Segmentation, Sign Recognition, CNN, Rescue Robot, RoboCup, GrabCut, Convex Hull

\footnote[0]{$^\dagger$ Corresponding Author: \href{mailto:m.rezaei@leeds.ac.uk}{m.rezaei@leeds.ac.uk} (M. Rezaei)}
\footnote[0]{$^\star$ Note: The first and second author have contributed equally}

\begin{figure}[t]
\centering
\includegraphics[width=0.95\columnwidth]{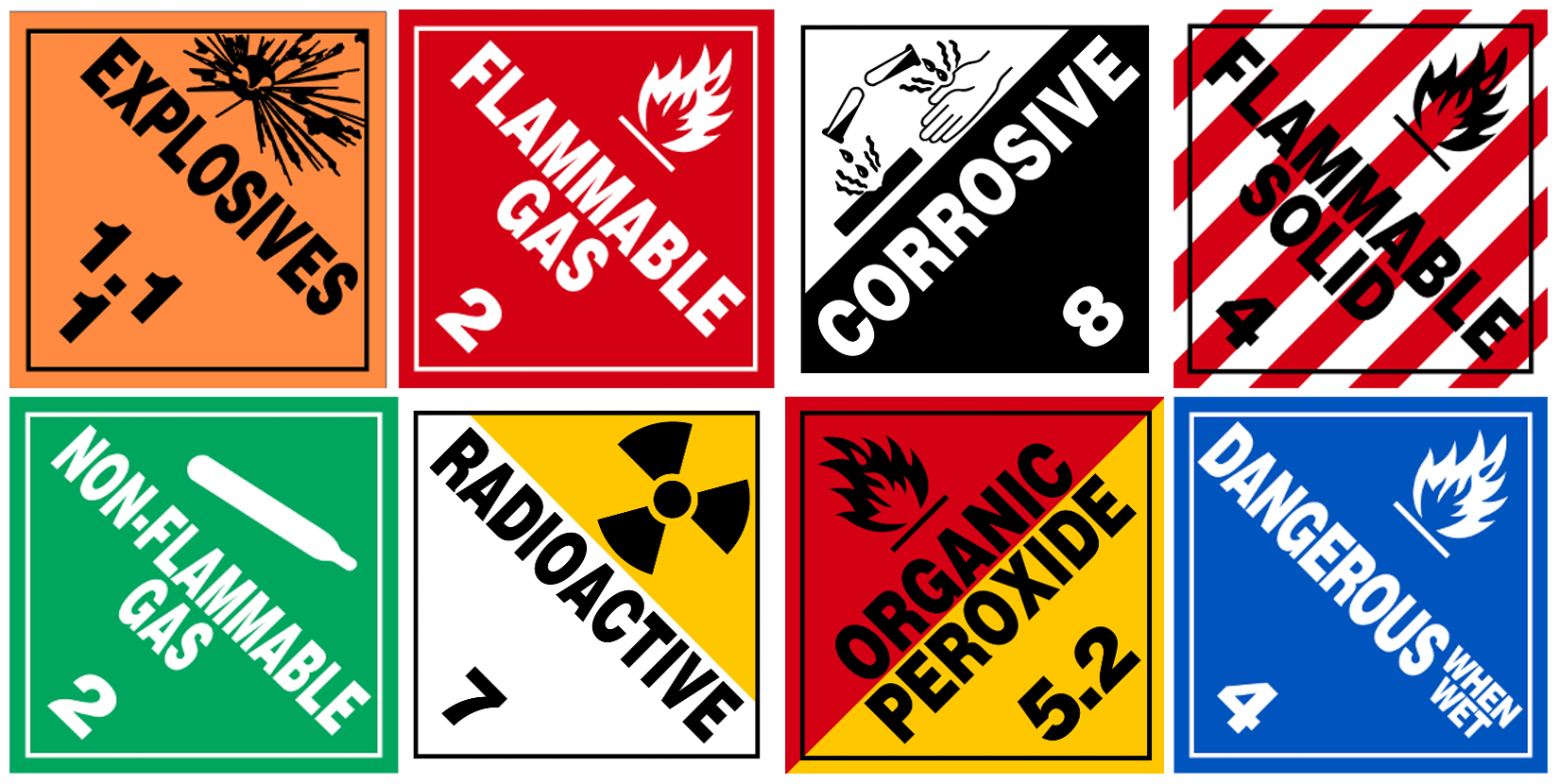}
\caption{Samples of standard and globally recognised Hazardous materials (HAZMAT) signs}
\label{sample-fig}
\end{figure}

\vspace{-3mm}
\section{Introduction}
\vspace{-1mm}
According to the U.S. Department of Transport (DOT) guidelines \cite{federal_hazmat}, all vehicles that transport dangerous goods or Hazmat materials/parcels must use large and clear Hazmat signs in front and back of the vehicles, to indicate the type of hazard for other road users. These signs identify information described by the sign shape, colour, symbols, and numbers. Figure \ref{sample-fig} shows some samples Hazmat signs \cite{inhs}. Because of the importance of Hazmat signs, all of them are globally accepted; therefore, each sign is language-independent. Every sign clearly describes the hazard type; so, the rescue team will be prepared to take action for the required tools and equipment to cope with the challenge.

The 2011 Fukushima Nuclear Power Plant disaster in Japan \cite{fukushima}, draw the attentions of lots of researchers to the importance of developing autonomous rescue robots for complex and dangerous tasks \cite{fukushima-robotics}. The recent development of Computer Vision and deep learning techniques has significantly helped to speed up the research \cite{rezbook2017}. In 2016, the RoboCup international committee, officially announced the Hazmat sign detection as one of the official competitions among rescue robot teams \cite{robocup}. Therefore HAZMAT sing-related research currently is in it's infancy with a very high research potential to discover the existing challenges and opportunities.

Hazmat detection in rescue scenarios is a challenging computer-vision based task due to environmental situations such as various lighting conditions, image perspective distortion, camera angles, image blurring, and frequent contrast changes.\\

A rescue robot has a limited power supply, memory and computational resources; and one the other hand, it needs to perform a real-time, accurate, and reliable sign detection mainly to save lives.  

Speed and accuracy are always contradictory to each other. i.e. developing a more accurate system normally decreases the speed, and vice versa. So we need to consider a trade-off in between, which makes the task even more difficult. 

This paper proposes a real-time method that provide a fast Hazmat sign detection, while maintaining a high level accuracy in various lighting conditions and different backgrounds. Furthermore, the system can be implemented on a small low weight mobile robots with limited memory, hardware and computational resources.

In the following sections we discuss more in-depth and provide further details. The rest of the paper is organised as follows: Section \ref{related-work} reviews the related work. In Section \ref{our-robot} we introduce the characteristics of our rescue robot which is developed to perform in real-world life saving scenarios. Then we provided the details of the proposed methodology in Section \ref{methodology} The outcome of the experimental results will be discussed in In Section \ref{experimental} and finally, Section \ref{conclusion} summarises the paper with concluding remarks.

\section{Related work}\label{related-work}
Recent approaches in sign detection can be categorised in five main classes:

\begin{enumerate}
\vspace{-2mm}
\item \textbf{Colour-based methods} mostly try to find the candidates' region of interest based on the colours of a Hazmat sign and 
then pass them to a conventional key feature extractor such as SIFT \cite{SIFT2004} or SURF \cite{SURF2008}.
 Based on research by Gossow et al. \cite{Surf-maching2008H}, the accuracy of colour-based techniques may significantly decrease at the distances of 1.5m and 2.0m to as low as 52\% and 20\%, respectively. Sensitivity to lighting condition and illumination changes are two other weaknesses of the colour-based methods.\\

\vspace{-2mm}
\item \textbf{Shape-based methods} try to find the candidate regions of interest using the appearance and shape characterise of each Hazmat shape. First, they generate the edge map of the region of interest (ROI) and then try to find the Hazmat attributes using shape-line attributes methods such as line or circle Hough transform \cite{shape-line}, \cite{shape-triangle-road}. These approaches may also find the candidate shape based on the SVM classifiers \cite{svm-road-sign} after an initial content recognition, or based on Gaussian-kernel SVMs. These methods are not occlusion-invariant, and also very sensitive to perspective distortions.\\

\vspace{-2mm}
\item \textbf{Saliency-based methods} aim to detect the silence objects, and then highlight and export them as the candidate regions. 
After that the accurate region of the sign will be detected using SIFT, SURF, ORB, Freak, or other detection methods\cite{hazmat-mobile2}, \cite{hazmat-mobile} \cite{saliency-model1998}, \cite{road-traffic-saliency}.\\

Most of the above-discussed technique have the following weaknesses in real-world scenarios:

\begin{enumerate}
\label{problems}
\item In cases where the Hazmat labels were arranged too close to each other, the detection process may completely fail.
\item The system fails to detect the hazmat signs in scenes with complex backgrounds.
\item The system only detects the hazmat signs from a very limited angles.
\end{enumerate}

\vspace{-2mm}
\item \textbf{Keypoint Matching Based Methods}. SURF is a keypoint detector that is invariant to image scaling and rotation which has been built based on the widely used SIFT detector, but SURF is considered to be much faster. Using integral images, the SURF algorithm applies average filters instead of the Gaussian filters in SIFT detector. The speed-up process plays an important role in the cases where the detector method must be fast enough to be considered real-time. The OpenCV's contribution modules provides some of the common keypoint detectors such as SIFT, SURF, and ORB which we evaluated them in this research. In order to detect Hazmats in real-time, every candidate was passed to the keypoint detector. Then a keypoint matcher between the keypoint database and the detected keypoints were used to recognise the object. The real-time performance of the algorithms was quite satisfying; however, the robustness of detection was an issue. Such classic methodologies simply could not lead to acceptable level of accuracy in our real-world and life-saving application.\\

\vspace{-2mm}
\item \textbf{Deep Learning based methods} 
Due to the weakness of the conventional object detection techniques \cite{sab2008a} as well as manually image engineering approaches, the research direction has redirected to deep learning based techniques. 
In deep learning based methods the models try to extract some common features of the hazmat sign during the training phase. 

In \cite{hazmat-surf-cnn-2019}, the authors aim to train a deep neural network model; however, the proposed model fails to detect hazmat signs in complex backgrounds.

Nils et al. \cite{CNN-uav}, train a deep neural network model based on the YOLOv2 algorithm for hazmat sign detection. Although their model performs fast on a GPU platform, their model is not real-time on CPUs and the system has an error of up to 1.5'' in localising the hazmat signs.

Another recent approach in this field is proposed in \cite{soren-cnn-2019}. In this method the system receives the visual and depth data of the environment by utilisation of an RGB-D camera.The proposed method computes the Homography to transfer between 3D and 2D perspective images. Then they rectify the image to cope with the distortion effects and the resulted images are feed to a CNN detector. Using the Homography matrix the system can also calculate the angle of the objects with respect to the camera coordinate system. The advantage of this approach is to detect the hazmat signs at various angles; yet it requires a very high computational cost. Therefore, this system is also not feasible for real-time performances. Furthermore, it needs additional sensors than common hazmat detection systems.

As a common weakness of the above methods, they only try to increase accuracy and do not consider the limited resources of a rescue robot which normally performs on an embedded CPU platform.
\end{enumerate}

Considering the reviewed research gaps and existing challenges, we will have three main contributions in this research. We publicly release a standard Hazmat dataset with PASCAL-VOC format as a new comprehensive dataset to be used by other researchers in the field. We introduce a CNN-based neural network model for Hazmat sign detection that successfully decreases the CPU usage by reducing the number of forwarded images into the network. Finally we develop a custom Non-Maximum Suppression (NMS) method that not only considers multiple bonding boxes and their confidence values, but also pays attention to the class of each bonding box before performing a blind suppression operation.

\section{Rescue Robots}\label{our-robot}

Before we dive in to the details of our methodology, we would like to provide some further information about the role of rescue robots and our designed ARKA rescue robot in Advanced Mobile Robotics Lab (AMRL). 
Rescue robots are designed to rescue people and/or
provide environmental data to the rescue team in order
to facilitate a rescue mission. The robots are mainly employed in extreme situations such as natural disasters, chemical/structural accidents, explosive detection, etc. 

A rescue robot is a type of robot that can enter in dangerous disaster scenes and carry out rescue tasks on behalf of a human. Earthquake scenes, chemical sites, collapsed buildings and towers due to fire or explosion, are few examples that may take place in a daily basis all around the world. 
One of the most important factors in rescue operations is to find and save victims, in time.

\begin{figure}[t]
\centering
\vspace{2mm}
\subfloat[\vspace{2mm} Main body structure \vspace{2mm}] {\includegraphics[width=0.97\columnwidth]{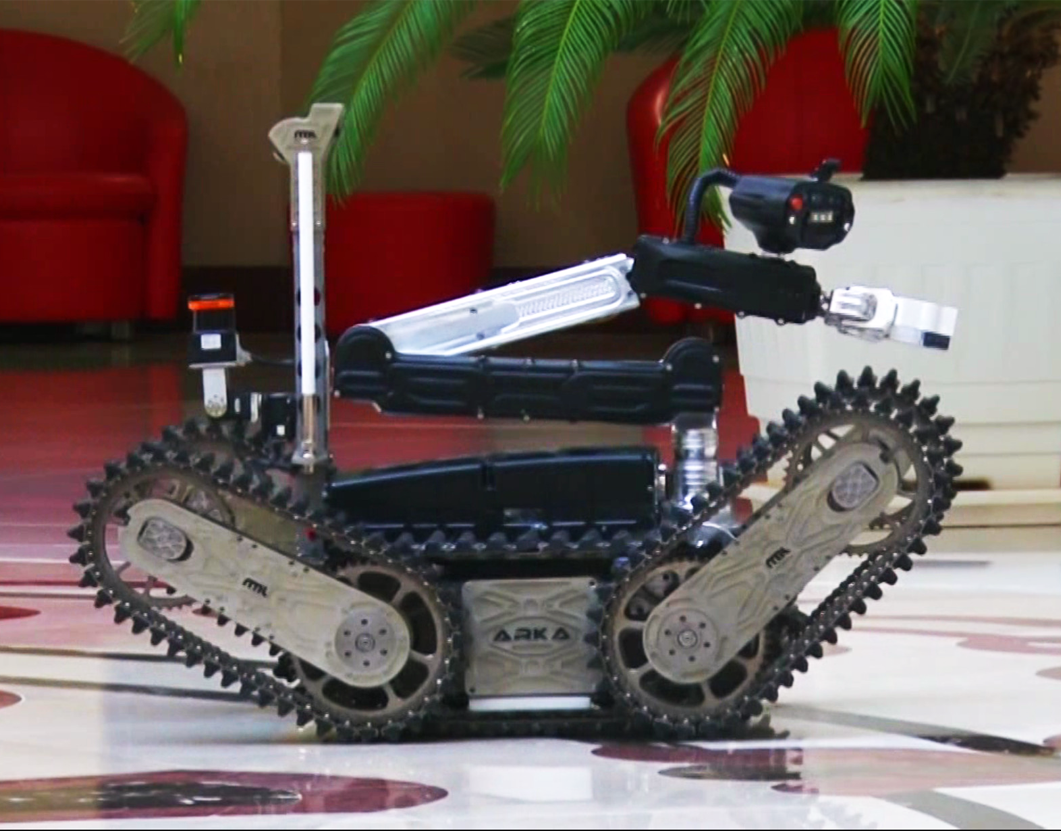}}\\
\subfloat[\vspace{2mm} Dexterous manipulator, 360° rotating wrist, gripper, microphone, depth and thermal cameras \vspace{2mm}]{\includegraphics[width=0.97\columnwidth]{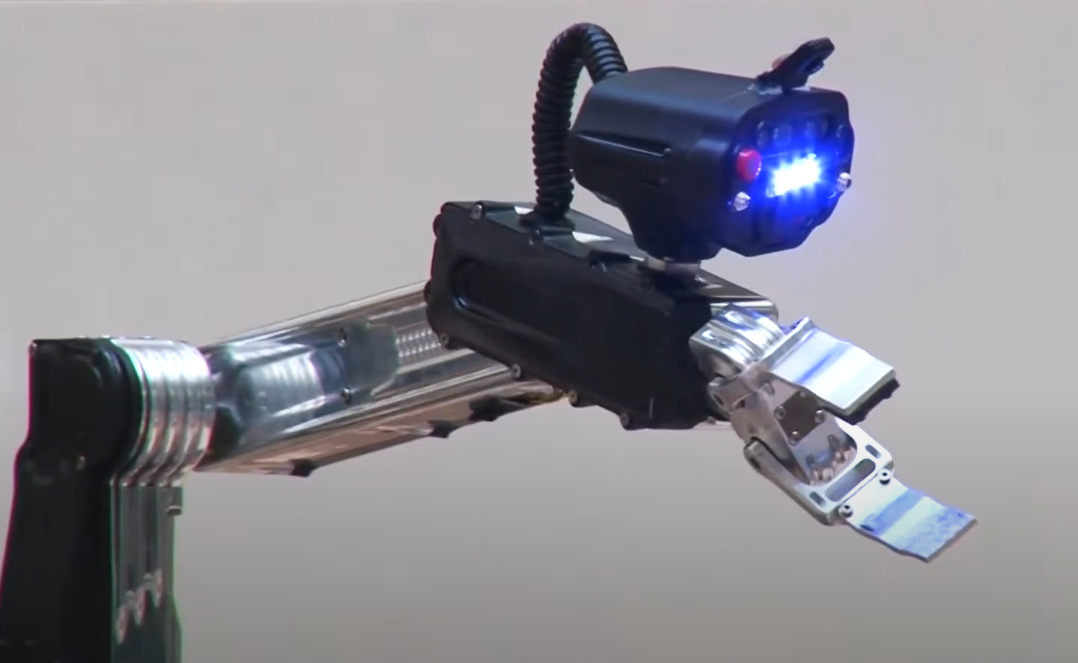}}
\caption{ARKA rescue robot, developed in our \href{https://youtu.be/i_7lMuh__iM}{Research lab}}
\label{arka-fig}
\end{figure}

\subsection{ARKA Rescue Robot}
ARKA is an unmanned ground vehicle (UGV) designed and developed by our team at Advanced Mobile Robotics Lab (Figure \ref{arka-fig}).
The ARKA is a transportable robot with superior mobility and advanced manipulation which is able to tackle dangerous situations. 
It weighs about ninety kilograms. It can climb stairs up to a gradient of 45° and slopes of 55°. The robot is equipped with a dexterous manipulator, 360° rotating wrist, gripper, microphone, depth, and thermal camera. The 13kg manipulator can be extended up to a length of 140cm. It can lift objects weighing more than 10kg at full arm extension and roughly 35kg at the close-in position.
The ARKA is suitable for missions such as explosive detection, explosive ordnance disposal (EOD) / bomb disposal, persistent observation, Gas leakage control, and vehicle inspections. The robot is very accurate and flexible in dealing with rescue and security problems, and it provides live video steams to map and localise in the unknown environments \cite{amrl}.

The robot is connected to an operator station via a 5GHz WLAN. We used two computing platform in our tests as follows: an Intel NUC mobile computer with an Intel Core i7-5557U processor for the robot, and another Intel NUC with a Core i5 processor for the remote operating station. No GPU is used on both sides.

The sensor box \cite{rez2010a} is a crucial and essential hardware part of tactical, rescuer, and police mobile robots to increase their perception capabilities. In our ARKA robot, this pack is equipped with various types of sensors and devices as follows: \\

\begin{itemize}
\vspace{-1mm}
\item \textbf {RGB-D Camera}: The Sensor Box is equipped with an Intel\textregistered RealSense\texttrademark \, depth-camera, which provides depth images and RGB images. One of the major usages of this sensor is to detect and avoid impassable grounds and obstacles, by using Point Clouds gained from the camera.
Furthermore to create a map of the current scene, the depth data is used for readiness tests in identification and dexterity operations. 

In order to detect any object in a rescue scenario, the main camera (Intel RealSense) is used to determine objects’ locations in the 3D environment as well as obtaining the relative distance of the detected Hazmat labels to the robot. Using the SLAM algorithm for self-localisation of the robot, this relative position can then be related to the created map.\\

\vspace{-1mm}
\item \textbf{CO2 sensor}: In order to find out whether the victim is breathing or not, an ``MQ-9'' sensor is being used which can be customised for detecting other types of gases upon request). \\

\vspace{-1mm}
\item \textbf {Thermal Camera}: One of the most important vital signs, for analysing whether the victim is still alive or not, is the temperature of the victim's body. Body position estimation and body temperature detection of the victim is accomplished by the combination of the previously discussed RGB-D sensor and equipping the autonomous robot with a Thermal Image sensor (miniAV160). This makes it capable of synchronous capturing of visual and thermal images. \\

\vspace{-1mm}
\item \textbf{Analog Cameras}: Two analog cameras that are mounted on the Sensor Box, assists the operator to drive the robot and accomplish particular missions. \\

\vspace{-1mm}
\item \textbf{Input and Output Audio}: A microphone and a speaker is installed on the Sensor Box to provide full-duplex audio communication between the victim and the rescue team, at any time if needed.\\

\vspace{-1mm}
\item \textbf{Laser Scanner}: is used for some autonomous tasks, were we need to calculate the distance between the robot and obstacle(s). Moreover, for generating a 2D map of the environment we use a Hokuyo UTM30-LX LIDAR attached to a stabiliser. This we guaranty that on sloping surfaces manoeuvres, the sensor stays parallel to the ground for a continuous mapping.
\end{itemize}

\section{Methodology}\label{methodology}
In this section we discuss four major steps that we have taken to develop our methodology based on an appropriate DNN model: A) Creating a training dataset, B) implementing a customised Non-maximal suppression function C) Data Feeding Optimisation, and D) Data Logging.

\subsection{Dataset}\label{dataset}
One of the major challenges in deep learning based methodologies is the requirement of large training dataset in order to achieve excellent results~\cite{dnn20}.

Convolutional Neural Networks or CNNs are supervised learning approaches, i.e. the labelled images that constitute as ground truth data must be initially provided to train the neural network. Preparing a good dataset is as important as a good neural network structure. Having no properly engineered dataset, it is very unlikely to get maximum performance out of a network.

We have developed a comprehensive dataset of Hazmat sign, including 1685 images in various angels, distortions, and different illumination conditions. The dataset has been divided into 13 different classes as per table \ref{hazmat-signs}.

\begin{table}[t]
\vspace{2mm}
\caption{13 categories of HAZMAT signs provided in the DeepHAZMAT dataset.}
\renewcommand\arraystretch{1.3}
\centering
\begin{tabular}{|l | l|}
\hline
\multirow{2}{*}{1- Poison}& \multirow{2}{*}{\includegraphics[width=0.09\linewidth]{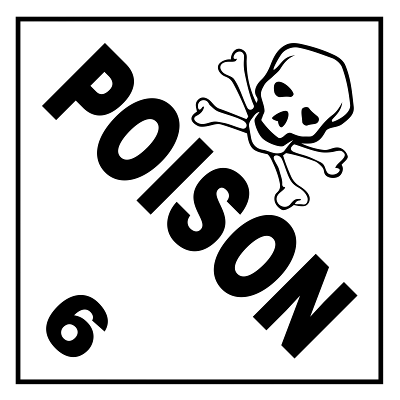}}\\ 
& \\ \hline
\multirow{2}{*}{2- Oxygen} & \multirow{2}{*}{\includegraphics[width=0.09\linewidth]{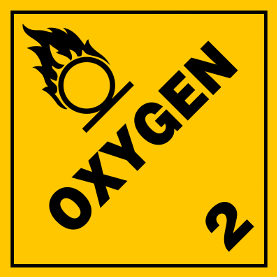}}\\ 
& \\ \hline
\multirow{2}{*}{3- Flammable Gas} & \multirow{2}{*}{\includegraphics[width=0.09\linewidth]{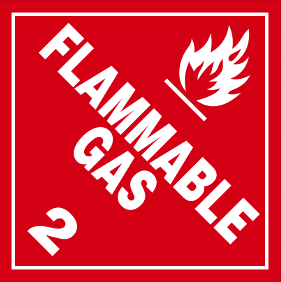}}\\ 
& \\ \hline
\multirow{2}{*}{4- Flammable Solid} & \multirow{2}{*}{\includegraphics[width=0.09\linewidth]{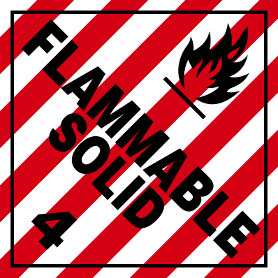}}\\ 
& \\ \hline
\multirow{2}{*}{5- Corrosive} & \multirow{2}{*}{\includegraphics[width=0.09\linewidth]{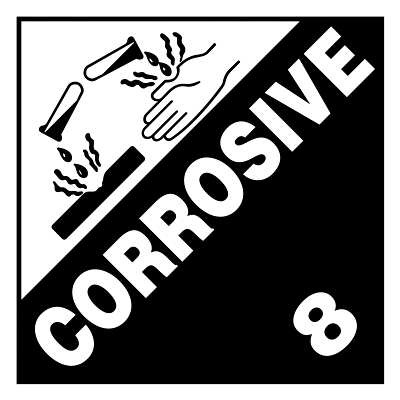}}\\ 
& \\ \hline
\multirow{2}{*}{6- Dangerous} & \multirow{2}{*}{\includegraphics[width=0.09\linewidth]{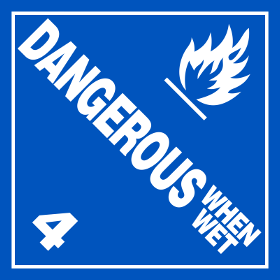}}\\ 
& \\ \hline
\multirow{2}{*}{7- Non-flammable Gas} & \multirow{2}{*}{\includegraphics[width=0.09\linewidth]{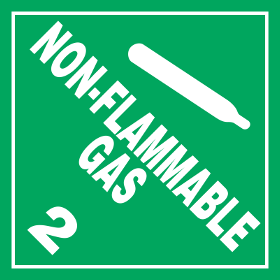}}\\ 
& \\ \hline
\multirow{2}{*}{8- Organic Peroxide} & \multirow{2}{*}{\includegraphics[width=0.09\linewidth]{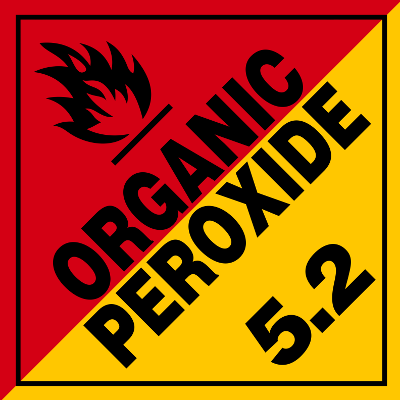}}\\ 
& \\ \hline
\multirow{2}{*}{9- Explosive} & \multirow{2}{*}{\includegraphics[width=0.09\linewidth]{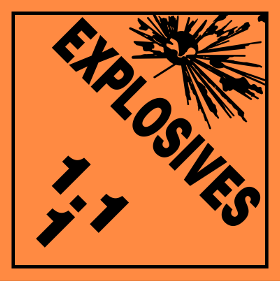}}\\ 
& \\ \hline
\multirow{2}{*}{10- Radioactive} & \multirow{2}{*}{\includegraphics[width=0.09\linewidth]{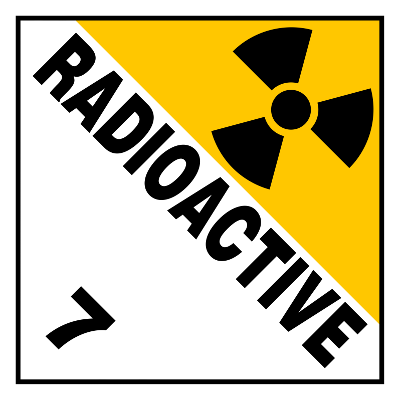}}\\ 
& \\ \hline
\multirow{2}{*}{11- Inhalation Hazard} & \multirow{2}{*}{\includegraphics[width=0.09\linewidth]{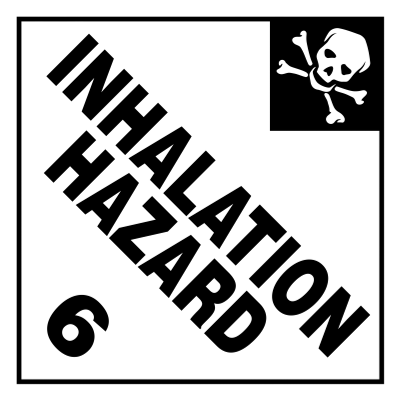}}\\ 
& \\ \hline
\multirow{2}{*}{12- Spontaneously Combustible} & \multirow{2}{*}{\includegraphics[width=0.09\linewidth]{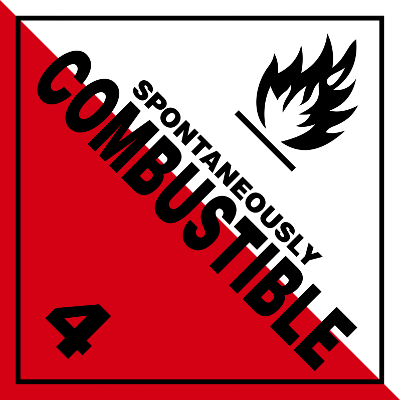}}\\ 
& \\ \hline
\multirow{2}{*}{13- Infectious Substance} & \multirow{2}{*}{\includegraphics[width=0.09\linewidth]{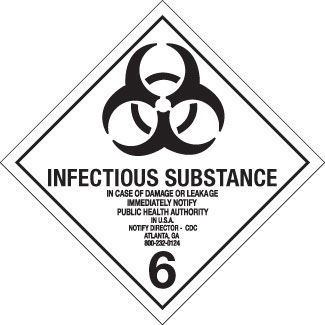}}\\ 
& \\ \hline
\end{tabular}
\label{hazmat-signs}
\end{table}

\begin{figure}[t]
\vspace{2mm}
\centering
\includegraphics[width=0.98\linewidth]{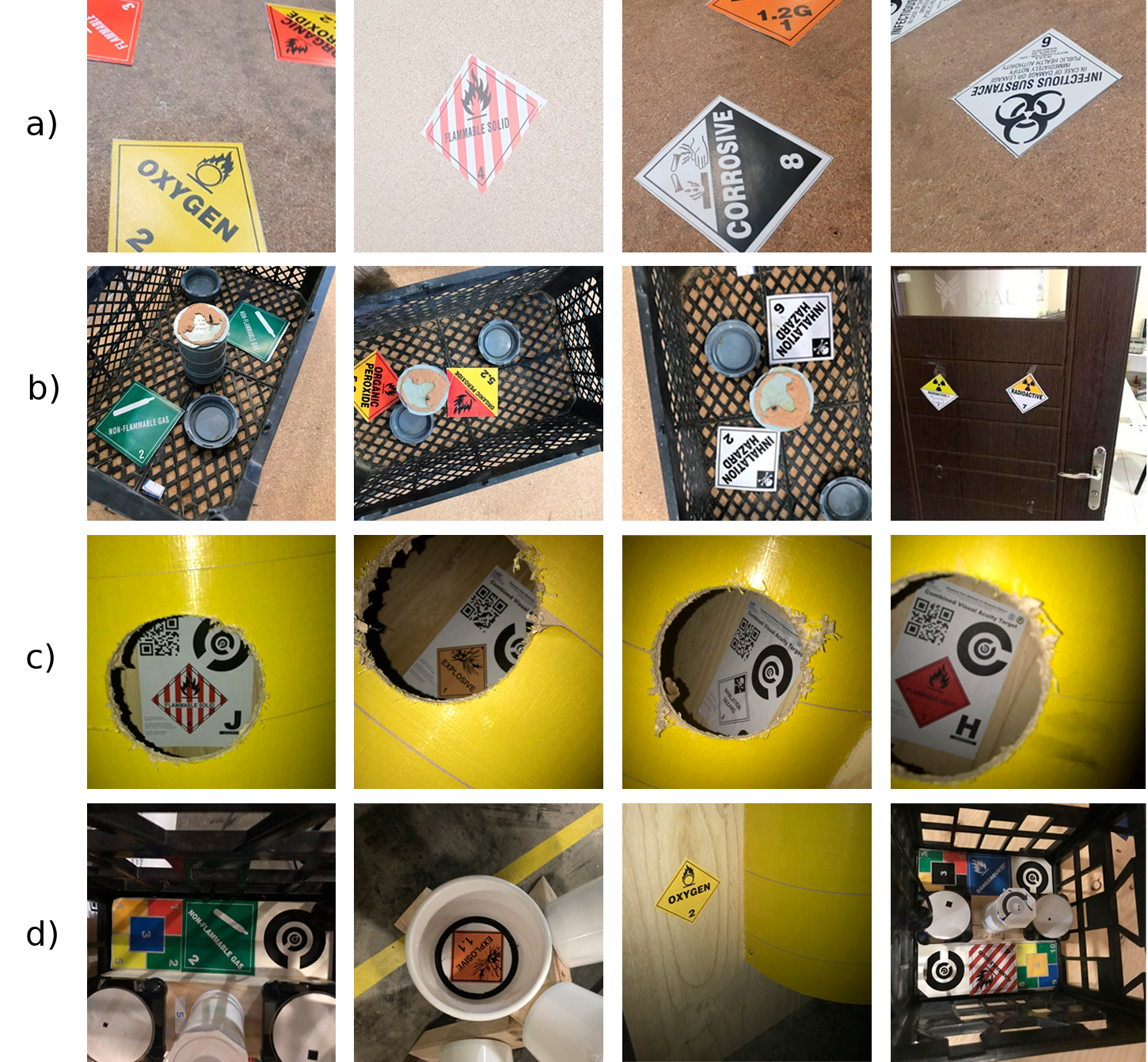}
\caption{ Row a, b: Samples in different angles, lighting conditions, and backgrounds. Row c , d: Samples in different situation at RoboCup competitions in past years.}
\label{fig-dataset}
\end{figure}

The dataset is annotated with PASCAL-VOC format as it is easy to convert into other annotation formats such as YOLO or COCO. Furthermore, the dataset can be easily labelled using the labelImg \cite{labelimg} labelling tool. 

The original dataset consists of 1685 image. Figure \ref{fig-dataset} shows some Hazmat sign samples from our developed dataset. 
However, we needed to increase  the number of hazmat signs to improve the performance of our algorithm. Besides, the dataset needs to be balanced and the number of images for each class should be almost same to have a Homogeneous dataset. 
Also, the size of our dataset should neither be very small that lead to model under-fitting and detection accuracy loss, and nor too large to increase the complexity of the feature extraction and overfitting challenges. 
To aim this, and using the augmentation technique, we expanded the dataset to more than 4000 images per class (4065) and in overall 52835 images that we split them into 80\% train and 20\% test dataset.

\begin{algorithm}[t]
\vspace{2mm}
\SetAlgoLined
\DontPrintSemicolon
 k := 5; \quad q := $2^k$\;
 p := q; \quad n := 0\;
 \While{hasNewFrame}{
  n = n + 1\;
  \If{n $>$ p}{
   n = 0\;
   frame := getFrame()\;
   objects := detectHazmats(frame)\;
   \eIf{len(objects) $>$ 0}{
    \If{p $>$ 1}{
        p = p / 2\;
    }
   }{
    \If{p $<$ q}{
        p = p * 2\;
    }
    }
   }
 }
 \caption{Increase/Decrease skip frames value}
\label{alg1}
\end{algorithm}


\begin{figure*}[t]
\centering
\includegraphics[width=1.02\linewidth]{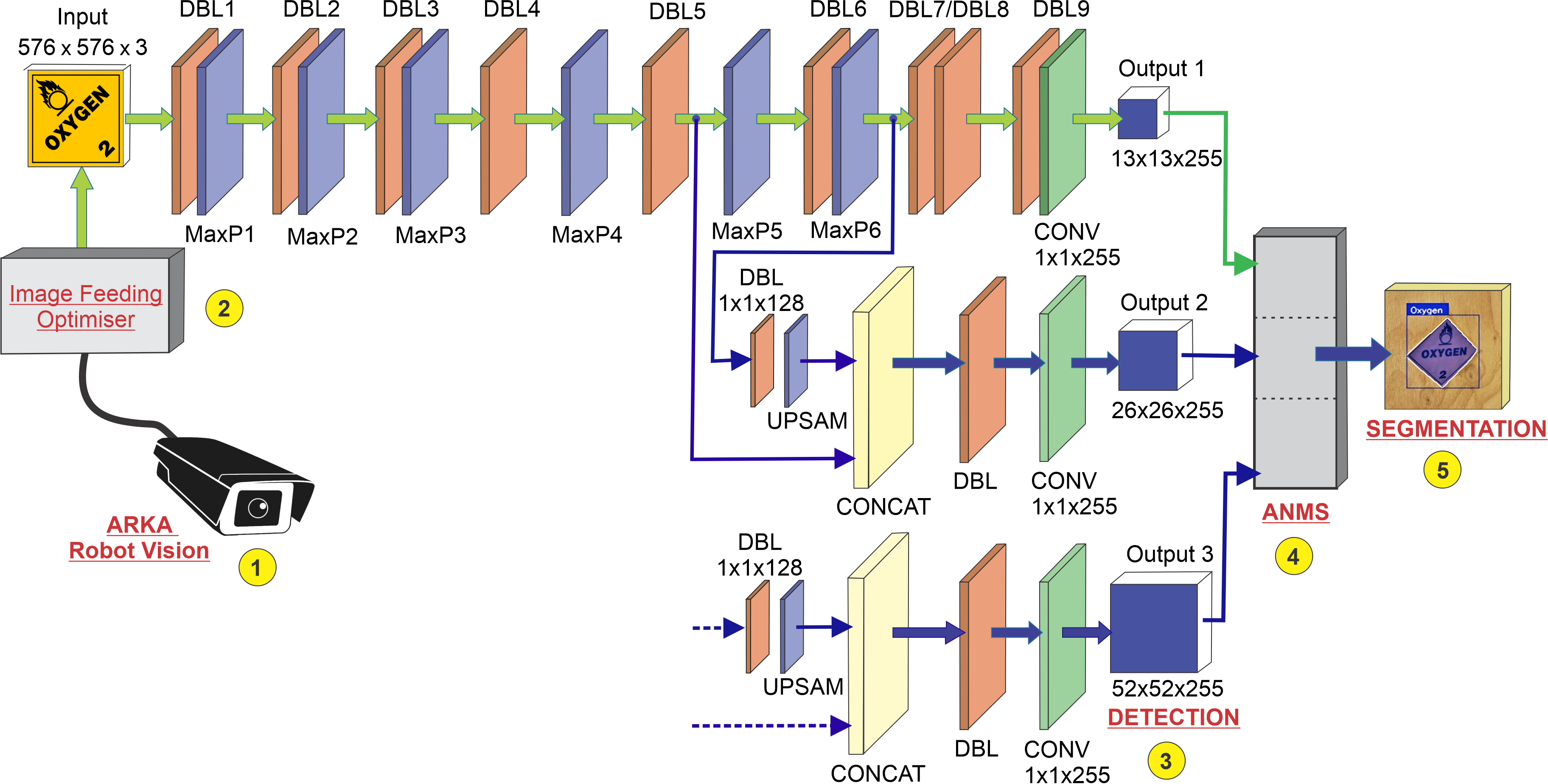}
\vspace{-2mm}
\caption{Structure of the five main modules of the developed HAZMAT recognition system including 1) ARKA Robot Vision module, 2) Image Feeding Optimiser, 3) YOLOv3-tiny based Hazmat detector, 4) Adaptive Non-maximal Suppression module, and 5) Segmentation module}
\label{yolov3}
\end{figure*}

For choosing our CNN model we had to consider a couple of more factors: 
\begin{enumerate}
\label{yolo3-table}
\item To be fast enough and implementable on mobile robots with restricted CPU resources,
\item To be accurate enough to get one of the best detection performances among state-of-the-arts.
\end{enumerate}

YOLO is one of the most common object detection networks, 
which can be appropriately reconfigured by
changing the size and the number of layers to satisfy a trade-off between speed and accuracy for our model. 
Since we are limited to use a low power CPU, we deploy a lighter version of YOLO, named YOLOv3-tiny \cite{yolov3}.
Figure \ref{yolov3} illustrates the architecture of our improved YOLOv3-tiny model. We provide with further details in the next three subsections. 

\subsection{Image Feeding Optimisation}

In a live video sequence the majority of the image frames may not include any hazmat signs, so it would be wise that we only look for the image frames which more likely contain a hazmat sign, rather than searching the entire frame sequences. This will significantly reduce the CPU usage. 
As a brief overview on our design, we perform a quick search for hazmat signs only within some of the input frames (for now, let's say on 50\% of the input frames depending on the camera frame rate) until we notice that some regions of an specific frame can be candidate regions of interest. Then we focus on more consecutive frames and those ROI to recognise and verify the hazmat sign and its type. 

\begin{figure*}[t]
\centering
\includegraphics[width=0.85\linewidth]{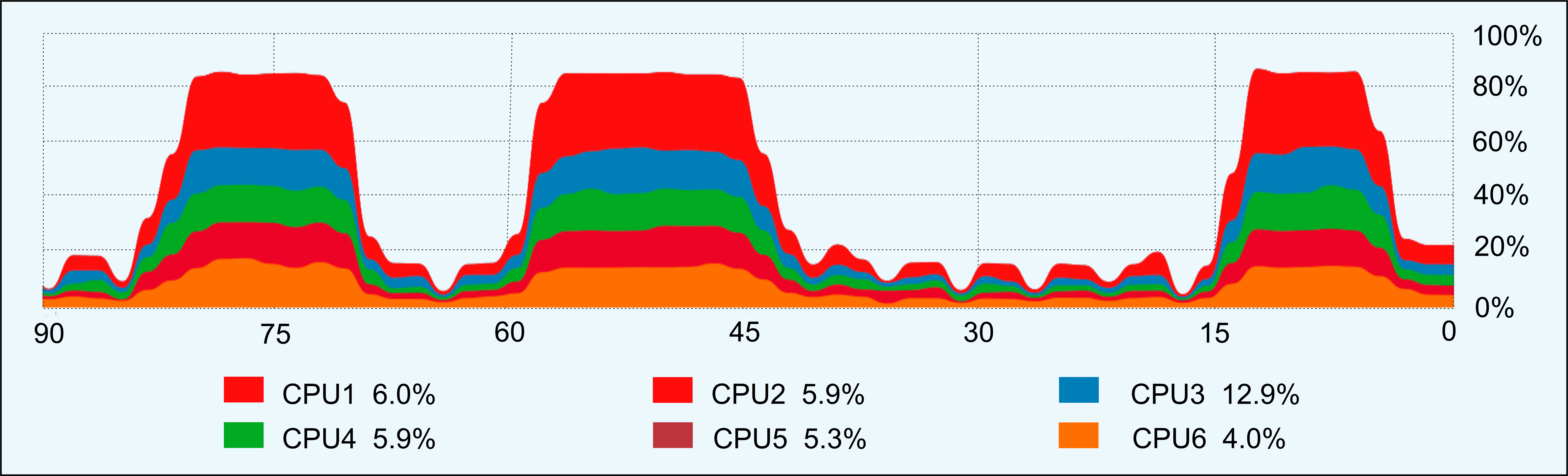}
\caption{A 90-Second sample graph of the 6-Core CPU usage for the proposed Image feeding optimisation}
\label{cpu_history_rgb}
\end{figure*}

Algorithm~\ref{alg1} provides further details about our approach. The action is similar to the way a human searches for recognising particular objects in an unknown environment. A human first scans the entire environment quickly, and then if during the scan process, his/her attention drawn to a particular region, he/she will focus on the ROI with more concentration and accurate search to recognise the object.

In Algorithm~\ref{alg1}, we have two main parameters $p$ and $q$. If we assume $S_c$ as the camera speed in frames per second ($fps$), then we set $q = 2^k$ where $k \in \mathbb{Z}$ and $q$ is the smallest squared integer number which is greater than $S_c$, so  that the quotient of the division operation $q \div S_c$ becomes greater than 1. For example, if the camera speed is 30$fps$ then $k$ would be equal to 5, because $q = 2^5 = 32$ and the quotient of $32 \div 30 = 1$. In other word, we initially only analyse one frame in every  second to find hazmat signs (1 frame out of every 32 frames).

In order to proceed with Image feeding optimisation, we initially process one frame for every $p$ frames; and as per the Algorithm~\ref{alg1}, at the beginning $p = q = 32$. If the system detects any kind of hazmat signs, we decrease $p$ by dividing it by 2 (i.e. $ p = p \div 2$) and again we process one frame per $p$ frames. As long as we repeatedly see hazmat signs in the input frames we halve the $p$ down to 16 then 8, 4, 2, and 1. Therefore, for every new time, $t$, we process a double number of frames than the previous time, $t-1$. As long as we see the hazmat signs, $p$ will keep decreasing until it reaches to 1. Otherwise, in case of no hazmat sign detection, $p$ will be doubled repeatedly until its maximum possible value (i.e. 32 in the above example). In other words, we only let the system analysis more images frames, and consequently more CPU usage, if the system guess a high chance of the hazmat signs in the current and upcoming frames.

Figure \ref{cpu_history_rgb} depicts the amount of CPU usage after applying the image feeding optimisation. 
The horizontal axis represents time for 90 seconds, and the vertical axis represents the percentage of the CPU usage. 
The diagram shows the usage of 6 processor cores of an Intel Core i7 CPU, in different colours. As can bee seen in Figure \ref{cpu_history_rgb}, there are some cases that we have very limited CPU usage (under 20\% in total). These are the instances where there has been no clue of the hazmat signs in the input images, and consequently minimum number of frames has been assessed per second. On the hand hand, there are three cases where the CPU usage has rapidly increased to above 82\% in just a few seconds. These are the cases that we have detected some ROI as potential hazmat singes, and consequently more and more input frames has been analysed. 

As per Figure \ref{cpu_history_rgb}, the average CPU core usage for a 90-second sample time-stamp is around 40\% while without input feeding optimisation, the CPU usage would be constantly around 82\%.

\subsection{Adaptive Non-Maximal Suppression (ANMS)}

Traditional object detection algorithms use a multi-scale sliding-window-based approach to search for a particular object in a window. Each window receives a score, depending on the number of matching features found inside the query window. Windows with a higher score than the set threshold will be marked as the candidate object regions. The final step of such approaches is to remove multiple neighbouring bonding boxes which point to the same instance of the object. This post-processing step is called non-maximal suppression (NMS).

In DNN-based object detection algorithms, the sliding-window approach is replaced with category independent region proposals using a CNN. Similarly a non-maximal suppression is also used in DNN based models to obtain the final set of detections. This significantly reduces the number of false positives \cite{soft-nms}.

Although occluded or overlapped signs may rarely appear in rescue operation scenes, we have to be cautious about it before suppressing them using a blind NMS approach. Standard NMS models significantly suppresses the overlapped bounding boxes, by keeping only the most confident ones and skipping the less confident bonding boxes. Non-Maximal Suppression also ensures that we would not have any redundant or extraneous bounding boxes. In some cases, the YOLO can detect partially overlapped objects and signs; however, it does not apply non-maximal suppression. Therefore, we would require to explicitly apply the NMS in our model. The standard NMS implementations (e.g. in OpenCV) does not care about the class of the occluded signs, and simply suppresses them all together. In contrast, we implement an adaptive version of Non-maximum suppression functions which we call it ANMS. The ANMS not only takes the class of the bounding boxes into account but also considers the confidence score of each bonding box to maintain the maximum benefit of the NMS without suppressing the important information.

As per the algorithm \ref{algo-anms}, the suppression process depends on a threshold value and the selection of threshold value is key parameter in performance of the model. As shown in the algorithm, instead of selecting the highest confidence value of a set of neighbouring bonding boxes, we select the highest confidence value of the same classes to make sure we do not suppress different classes even with a lower confidence levels.

\begin{algorithm}[t!]
\SetAlgoLined
\DontPrintSemicolon
\caption{Adaptive Non-Maximal Suppression}
\SetKwFunction{FMain}{ANMS}
    \SetKwProg{Fn}{Function}{:}{}
    \Fn{\FMain{$B,S,C,t$}}
{
    $D \leftarrow \{\}$\;
    \While{$B \neq$ empty}{
        m $\leftarrow$ selectMaximumConfidence($S$, $C$)\;
        $M \leftarrow b_m$\;
        $D \leftarrow D \cup M$\;
        $B \leftarrow B - M$\;
        \For{$b_i$ in $B$}{
            \If{ IoU ($M$, $b_i$) $\geq$ $t$}{
                $B \leftarrow B - b_i$\; 
                $S \leftarrow S - s_i$\; 
                $C \leftarrow C - c_i$\;
            }
        }
    }
    \Return{$D, S, C$}\;
}
\textbf{End Function}
\label{algo-anms}
\end{algorithm}

In Algorithm \ref{algo-anms}, $B_i$ is the list of initial detect bonding boxes, $D_j$ is the list of final detections, $S_i$ represents the corresponding detection scores, $C_k$ contains corresponding detection classes, and $t$ is the NMS threshold.
Non-maximum suppression starts with a list of detection boxes $B$ with scores $S$ and classes $C$. After choosing the bonding box with the maximum score $m$ in the class $C$, the remaining bounding boxes from the same class will be removed from the set $B$ and the appends it to the $D$. It also removes any box which has an overlap greater than a threshold $t$ with $M$ in the set $B$. This process is repeated for remaining boxes $B$ and classes .

The NMS algorithm that we use in ANMS is based on the Blazing Fast-NMS developed by Tomasz Malisiewicz \cite{tomas2011} which is over 100x faster than older NMS algorithms.

\begin{figure}[t]
\centering
\includegraphics[width=0.97\columnwidth]{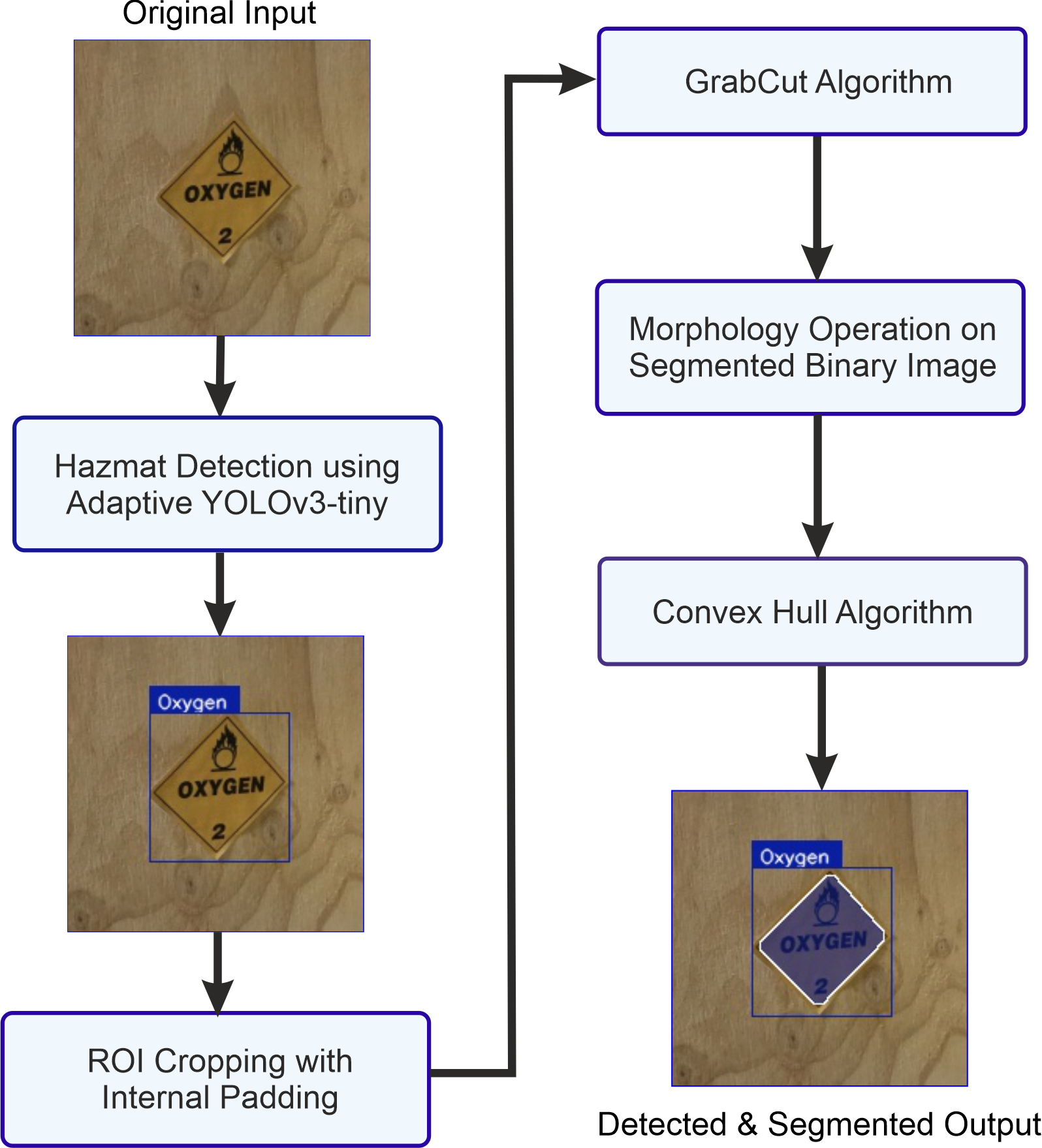}\\
\vspace{3mm}
\caption{Flowchart of the proposed method}
\label{GrabCut-chart-fig}
\end{figure}

\begin{figure*}[h!]
\centering
\includegraphics[width=0.9\linewidth]{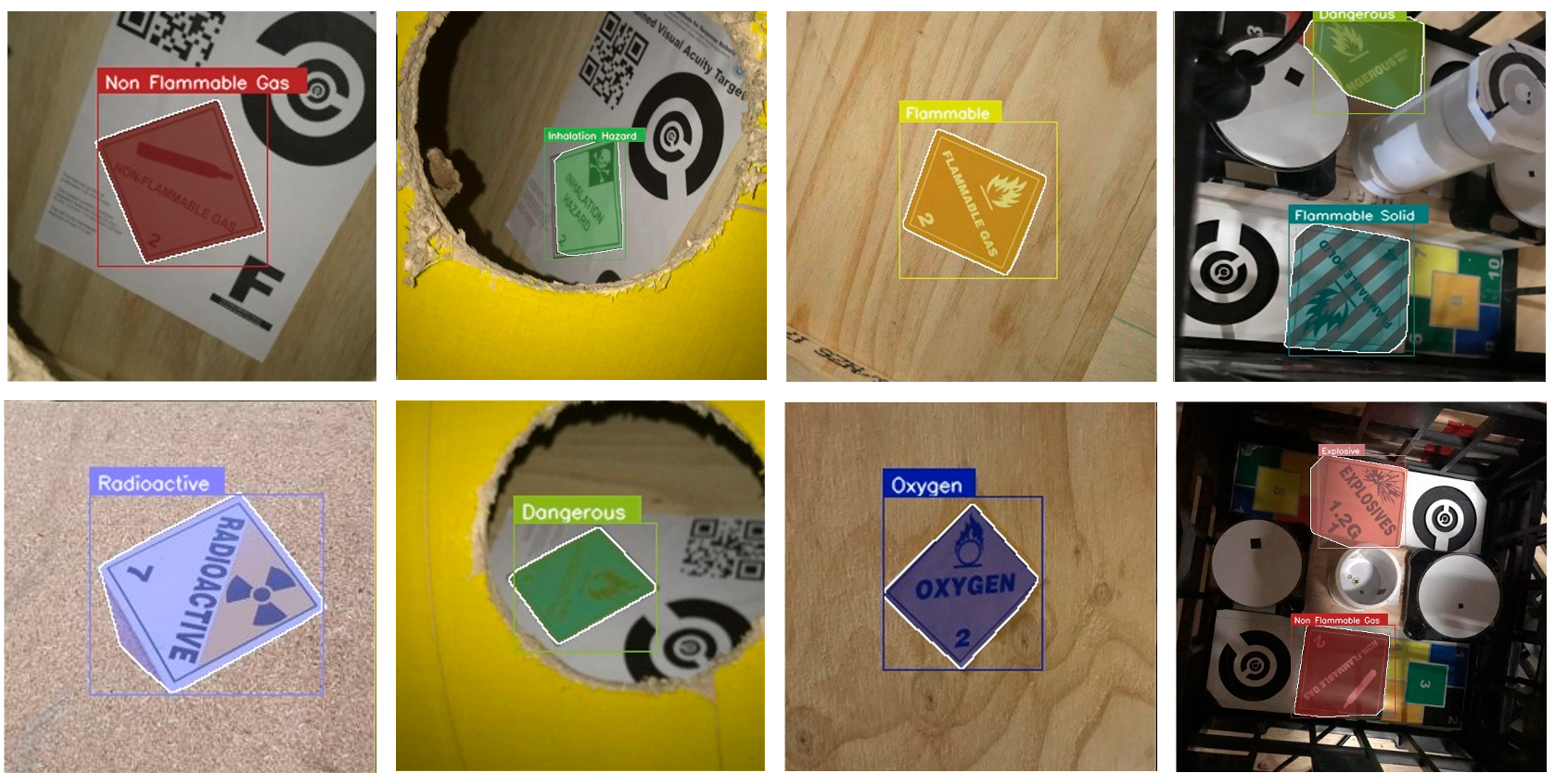}
\vspace{2mm}
\caption{Overall view of HAZMAT detection and segmentation, using our adaptive YOLOv3-tiny and GrabCut Segmentation}
\vspace{2mm}
\label{hazmat_segmentation}
\end{figure*}

\subsection{Hazmat Sign Segmentation}
YOLO bonding box outputs are in the form of upright rectangles; however, the predicted hazmat signs may have been rotated. This may cause part of the background segments also appears in the Hazmat sign bonding boxes (see Figure~\ref{GrabCut-chart-fig}).

To cope with this, background removal methods can been used to separate the detected hazmat sign from the background. Such operations help to have a more accurate position of the object and its approximate angle. In some environments, the hazmat sign may have a colour match with the background or other objects next to the detected sign, and the algorithm may consider that these objects and colours are part of the sign, and ultimately make a mistake in separating the sign from the background. To solve this problem, it is recommended to use edge detection and noise elimination algorithms.

GrabCut \cite{GrabCut-org} is an image segmentation method, based on iterative graph cuts as in \cite{GrabCut1}, \cite{GrabCut2}, \cite{GrabCut-leaf}, \cite{GrabCut-humans}. The algorithm estimates the colour distribution of the target object and that of the background using a Gaussian mixture model. In GrabCut based segmentation algorithm, we pass our region of interest in form of a bounding box to extract the foreground (i.e. the Hazmat sign) from the background. Since our Neural network already provides a rather accurate bonding box for the Hazmat region, and the hazmat signs are colour-coded signs, we believe this will be an appropriate approach to take the advantage of using an adapted GrabCut technique for a dimension-independent and high-resolutions segmentation. 

As one of the very recent research works, but in a different application, \"{U}nver and Ayan \cite{skin-lesion} also use GrabCut technique for lesion skin segmentation. To the best of our knowledge no research has been performed in utilisation and adaptation of the GrabCut technique for Hazmat sign segmentation. In the next section (experimental results) we demonstrate a very high rate of Intersection over Union (IoU) achieved, in comparison with the the ground truth data and other conventional metrics.

We pass the YOLO bonding boxes outputs to the adapted GrabCut algorithm by applying a small $(\cong 5\%)$ internal padding.
Figure \ref{GrabCut-chart-fig} illustrates a flowchart of segmentation method. The final result includes two types of pixel-wise segments: Hazmat and non-hazmat segments. 

After that, we uses Convex Hull to gain a more accurate polygon segmentation that encompasses the hazmat sign boundary (See Figure \ref{hazmat_segmentation}). The convex hull for a set of pixel points $S$ in $n$ dimensions is the intersection of all convex sets containing $S$. For N pixel points $p_i = p_1, ..., p_N$, the convex hull $C$ is given by the following expression:

\begin{equation*}
 C = \left[ \sum_{j = 1}^N  \lambda_jp_j: \lambda_j> = 0 \mbox{ for all } j \mbox{ and }  \sum_{j=1}^N \lambda_j=1  \right]
\end{equation*}

\noindent were $\lambda_j = l_i/L$, and $l_i$ is the ratio of the length of each convex edge $i$ to $n$ to the total length of all edges $L=\sum_{j=1}^n l_j$.

\subsection{Data Logging and Dataset Expansion}
Collection and preparation of a large and multi-faceted dataset has always been a challenge in training deep-learning based models. Having a larger dataset leads to a better training and higher accuracy. We have created an event logger for the detection service that captures and saves the hazmat images during the real-world operations of the rescue robot, to create a secondary hazmat dataset. The collected hazmat signs can either be feed forwarded to the network for further training or to be saved and annotated later by an expert to expand the main dataset. This creates a more comprehensive train set for further development of  the model, hence, more accurate rescue operations in the future.

\section{Experimental Results}\label{experimental}

We trained out developed model on a PC platform, equipped with an Intel Core i7-6700 CPU, NVIDIA GeForce GTX 1080Ti GPU, 8 GB of Memory, and Ubuntu 18.04 OS.

\begin{figure*}[t!]
\centering
\renewcommand{\arraystretch}{1.1} 
\includegraphics[width=0.83\linewidth]{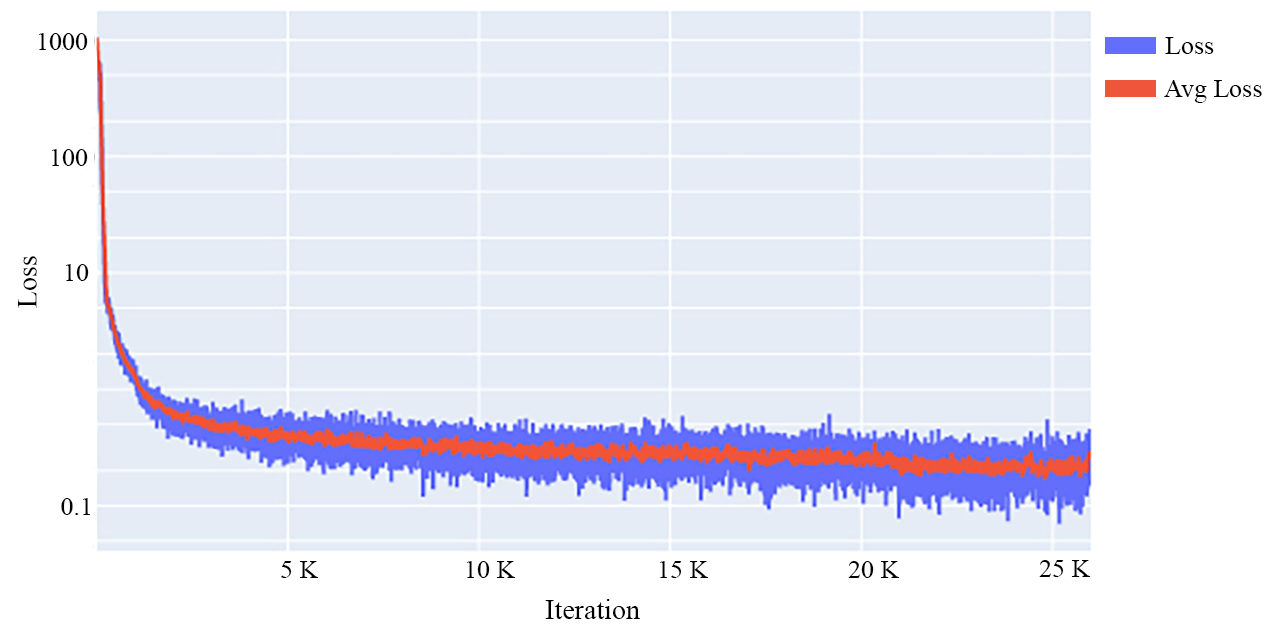}
\vspace{-1mm}
\caption{The training graph (Loss and Average Loss improvements over time and iterations)}
\vspace{4mm}
\label{train-yolov3}
\end{figure*}

To configure our custom YOLO model, we had to consider a trade-off between speed and accuracy. After several experiments, we find the best size of the input image for our model as $576\times 576$. Table~\ref{yolov3-tiny-config} provides the details of our custom training setting for the YOLOv3-tiny deep neural network. We set the learning rate as 0.001 with the batch size of 64. 
In the final stages of iterations in the training phase (between the iteration numbers 20800 and 23400), we multiplied the learning rate to 0.1 to make it smaller and proceed with a more precise weight adjustments, to prevent overfitting. 
Figure \ref{train-yolov3} shows the loss and average loss value of the model during the training phase, and after $25K$ iterations. 


\begin{table}[h!]
\centering
\caption{Training configuration of the YOLOv3-tiny for our hazmat detection robot}
\begin{tabular}{l|l}
\hline
\multirow{2}{*}{\textbf{Parameter}} & \multirow{2}{*}{\textbf{Value}} \\
& \\ \hline
Batch Size      & 64             \\ 
Subdivisions    & 16             \\ 
Momentum        & 0.9            \\ 
Decay           & 0.0005         \\ 
Burn In         & 1000           \\ 
Learning Rate   & 0.001          \\ 
Max Batches     & 26000          \\ \hline
\end{tabular}
\label{yolov3-tiny-config}
\end{table}

In order to assess the robustness of our method, we conducted different evaluation metrics to analyse the average precision of each classes. We used 80\% of the dataset for the train phase and the rest of 20\% unseen samples as the test dataset. 

We assessed the robustness of the algorithm against five metrics with the following results: precision rate = 94\%, recall-rate = 98\%, F1-score = 96, and Intersection over Union (IoU) = 81.83\% and mAP = 99.03\%.\\

\begin{itemize}
    \item \textbf{Precision rate:} $$\frac{\sum True Positives}{\sum (True Positives + False Positives)}$$
    \item \textbf{Recall rate:} $$\frac{\sum True Positives}{\sum (True Positives + False Negatives)}$$  
    \item \textbf{F1-score: } $$2 \times e\left (\frac{Precision \times Recall }{Precision + Recall} \right )$$  
    \item \textbf {Average Intersection over Union:} $$IoU = \frac{Summed Area Of Overlap}{Area Of Union} $$
		\item \textbf {Mean Average Precision \cite{mscoco}:} $$mAP = 99.03\%$$
\end{itemize}

\begin{figure*}[t!]
\centering
\renewcommand{\arraystretch}{1.1} 
\includegraphics[width=0.87\linewidth]{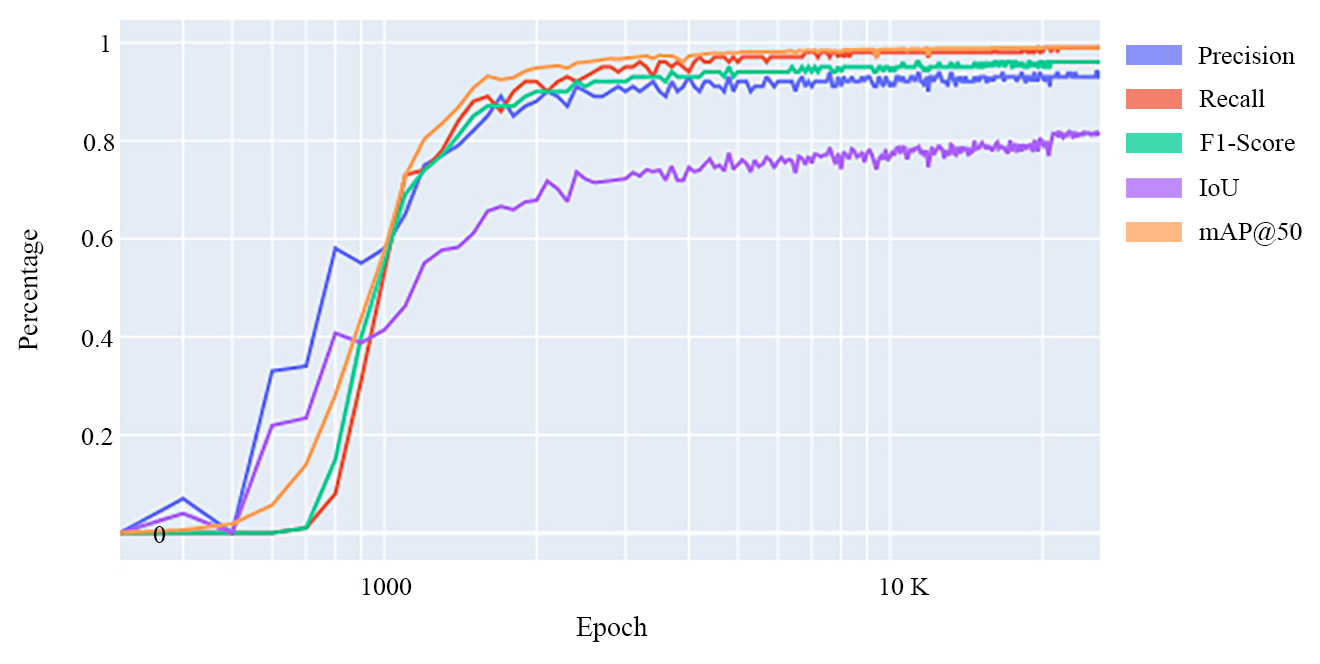}
\vspace{-1mm}
\caption{The training graph (Metrics vs. iteration)}
\vspace{1mm}
\label{train-metrics-yolov3}
\end{figure*}

\begin{figure*}[t!]
\vspace{2mm}
\centering
\renewcommand{\arraystretch}{1.1} 
\includegraphics[width=1\linewidth]{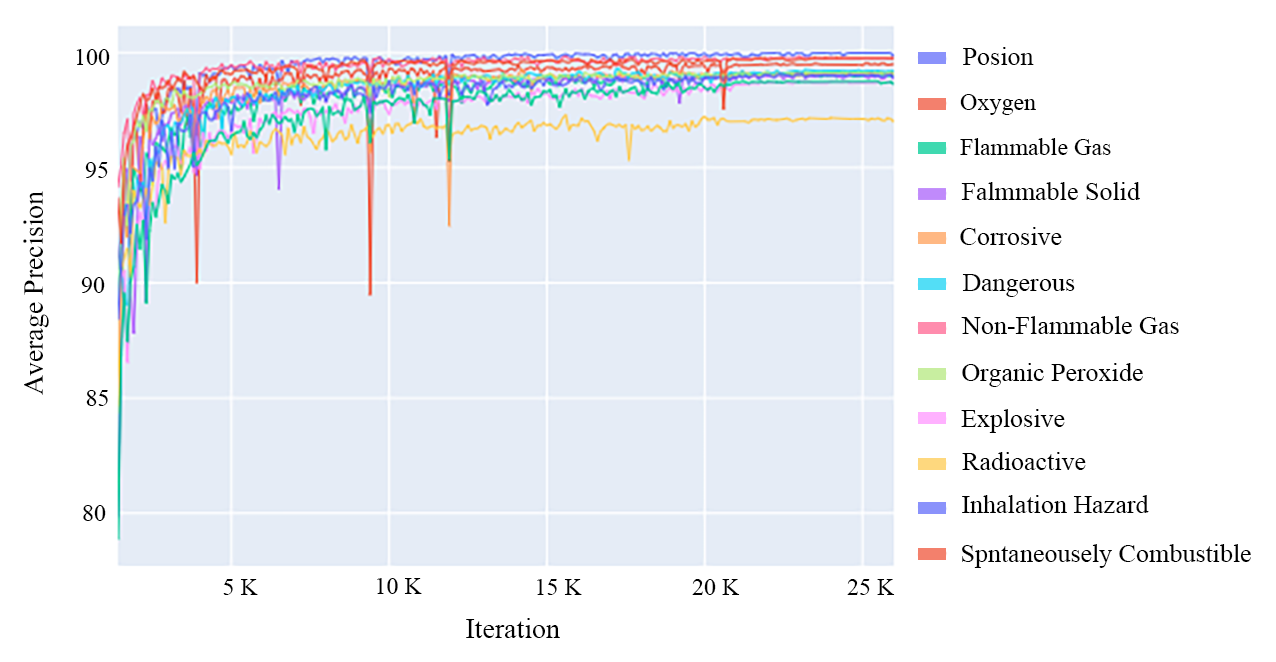}
\caption{The training graph (Each Class AP vs. iteration)}
\vspace{4mm}
\label{train-each-class-yolov3}
\end{figure*}

Figure \ref{train-metrics-yolov3} shows the gradual improvement of the precision rate, Recall rate, F1-score, Average IoU, and mAP after 25,000 Epochs. Figure \ref{train-each-class-yolov3} demonstrates the average precision rate of the proposed DeepHAZMAT model for all of the thirteen HAZMAT signs discussed in Section \ref{dataset}. 

Based on the visual appearance of the graphs shown in Figurers \ref{train-yolov3}--\ref{train-each-class-yolov3}, it can be also  confirmed that the model is not suffering from overfitting issue, so we can conclude that the hyperparameter tuning of the system has been successful and the system is robust and reliable enough in dealing with all of 13 discussed hazardous materials signs. 

In order to provide further information, we also considered the mAP (mean average precision) metric as another standard evaluation matrices proposed by MSCOCO \cite{mscoco} where AP is the average precision over multiple Intersection over Union (IoU). In our neural network with IOU-threshold of 50\%, the mean average precision (mAP@50) is equal to 99.03\%.

\begin{figure}[t]
\begin{tabular}{cccc}
{\includegraphics[width = 0.46\columnwidth]{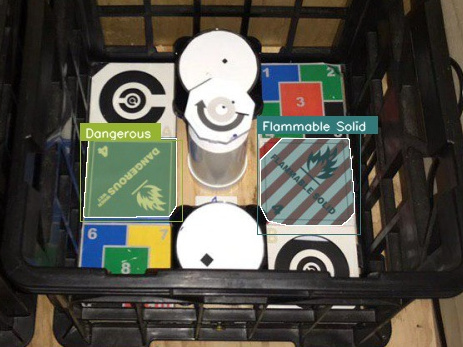}} & \hspace{-3mm}
{\includegraphics[width = 0.46\columnwidth]{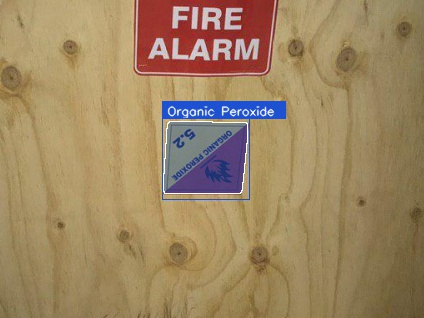}} \vspace{2mm} \\
\vspace{4mm}
{\includegraphics[width = 0.46\columnwidth]{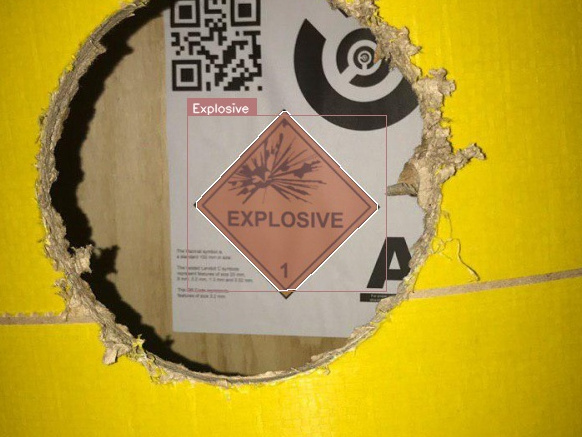}} & \hspace{-4mm}
{\includegraphics[width = 0.46\columnwidth]{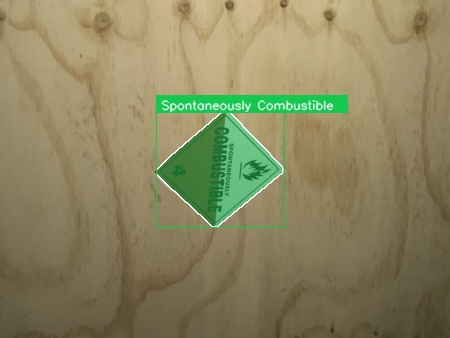}}
\end{tabular}
\vspace{-2mm}
\caption{Successful detection of Hazmat signs by ARKA robot on a challenging test field in Sydney, Australia}
\label{group_of_hazmats}
\end{figure}

\begin{figure}[t]
\centering
\includegraphics[width=0.95\columnwidth]{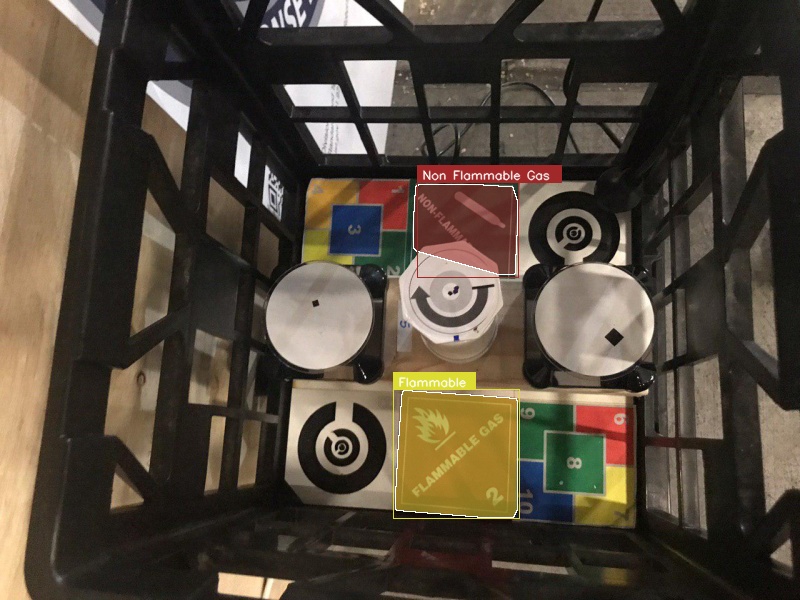}
\vspace{4mm}
\caption{Examples of signs detection and segmentation in challenging lighting condition with significant occlusions}
\label{occluded_hazmats}
\end{figure}

By providing figures \ref{group_of_hazmats} and \ref{occluded_hazmats} we would like to reiterate one the robustness and the performance of the system in two other challenging scenarios: Hazmat sign detection in complex backgrounds and lighting conditions as well as detection of occluded or partially visible Hazmat signs in complex real-world scenes and backgrounds, performed by ARKA robot. 

Table \ref{tab:result-table} shows the performance of the model for every classes using six evaluations metrics including Average Precision (AP), Precision Rate (PR), Recall Rate (RR), Accuracy (ACC), F1-Score, and Intersection over Union (IoU). Paying attention to the green and red numbers in each column of the table (as the best and weakest  performances, respectively), it can bee seen that while the algorithm is very robust in dealing with all 13 categories of HAZMAT signs, it performs slightly better for the Poison Hazmat sign detection and slightly weaker in Radioactive sign detection. This mean we probably need to add more diversity of sample radioactive hazmat signs to the training set. This will lead to gain a more balanced performance for all signs. 

\begin{figure*}[t!]
\centering
\vspace{4mm}
\includegraphics[width=0.95\linewidth]{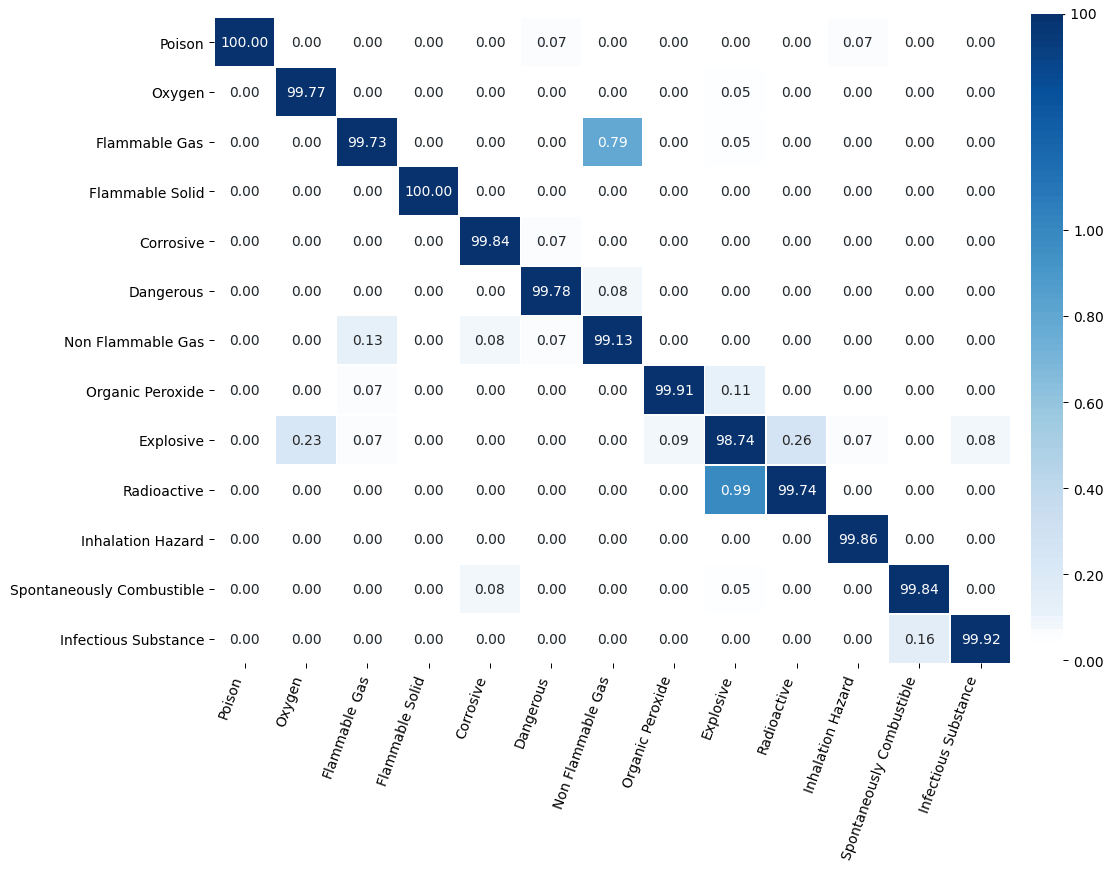}
\caption{Confusion Matrix of the 13 Hazmat Signs Based on the Precision Rate}
\label{confusion_matrix}
\end{figure*}

Figures \ref{confusion_matrix} provides a normalised confusion matrix to visualise the accuracy of the proposed method for every class. The horizontal axis demonstrate the actual labels and the vertical axis shows the predicted labels. The confusion matrix shows there are very limited instances where the DeepHAZMAT model may confuse the explosive and radioactive signs, interchangeably. A similar misclassification can be seen for Non-flammable gas signs, otherwise we can see a nearly perfect classification results based one the matrix diagonal. 

In Table \ref{tab:comparison} we compare the performance of the proposed DeepHAZMAT methodology with four other models (two classic models and two DNN-based models). We evaluated each method against nine metrics including five accuracy related metrics, three versatility based features, and finally, the overall speed of the model. 

As the table represents, none of the evaluated models are a definite winner in all metrics; However, DeepHAZMAT performs as the top one in five major metrics including recall rate, speed, segmentation, ability of multiple Hazmat sign detection, and adaptive bonding box feature. The proposed model also stays as the second best in terms of other accuracy metrics. 

The proposed DeepHAZMAT system only performs less than 1\% weaker in non-winning metrics which is negligible comparing to other important outperforming features, as well as extra capabilities of the model.  
Considering the main objectives of this research, which was accurate multiple HAZMAT sign detection, in challenging lighting conditions and environments, with restricted computational resources, the model well supports our requirements.

\begin{table*}[t!]
\centering
\caption{The experimental results and metrics for the accuracy of each Hazmat class}
\label{tab:result-table}
\renewcommand{\arraystretch}{1.1} 
\begin{tabular}{l|c|c|c|c|c|c}
\hline
\textbf{Class}            & \textbf{AP} \% & \textbf{PR} \% & \textbf{RR} \% & \textbf{ACC} \% & \textbf{F1-Score} \% & \textbf{IoU} \% \\ \hline
Poison                    & \textcolor[rgb]{0.2,0.8,0.2}{99.96} & 97.62 & \textcolor[rgb]{0.2,0.8,0.2}{100.00} & 99.80 & \textcolor[rgb]{0.2,0.8,0.2}{98.78} & \textcolor[rgb]{0.2,0.8,0.2}{89.32} \\ 
Oxygen                    & 99.78 & 97.63 & 98.76 & 99.72 & 98.19 & 88.10 \\ 
Flammable                 & 99.18 & 96.46 & 97.08 & 99.42 & 96.77 & \textcolor[rgb]{1,0,0}{85.91} \\ 
Flammable-solid           & 99.05 & 98.19 & 96.95 & 99.57 & 97.57 & 86.95 \\ 
Corrosive                 & 99.47 & 98.39 & 98.05 & 99.73 & 98.24 & 86.95 \\ 
Dangerous                 & 99.58 & 96.20 & 98.75 & 99.52 & 97.46 & 87.56 \\ 
Non-flammable-gas         & 99.90 & \textcolor[rgb]{0.2,0.8,0.2}{98.66} & 98.90 & \textcolor[rgb]{0.2,0.8,0.2}{99.82} & 98.77 & 88.53 \\ 
Organic-peroxide          & 99.34 & 98.51 & 98.51 & 99.80 & 98.51 & 88.84 \\ 
Explosive                 & 99.08 & \textcolor[rgb]{1,0,0}{95.02} & 97.66 & 99.21 & 96.32 & 87.78 \\ 
Radioactive               & \textcolor[rgb]{1,0,0}{98.40} & 95.58 & \textcolor[rgb]{1,0,0}{95.82} & \textcolor[rgb]{1,0,0}{99.20} & \textcolor[rgb]{1,0,0}{95.70} & 86.25 \\ 
Inhalation-hazard         & 99.27 & 95.97 & 97.80 & 99.48 & 96.88 & 87.43 \\ 
Spontaneously-combustible & 99.64 & 97.99 & 99.34 & 99.80 & 98.66 & 88.36 \\ 
Infectious-substance      & 99.20 & 95.85 & 97.55 & 99.47 & 96.69 & 87.50 \\ \hline
\textbf{All class metrics average}& \textbf{99.37} & \textbf{97.09} & \textbf{98.09} & \textbf{99.58} & \textbf{97.58} & \textbf{87.65} \\ \hline
\end{tabular} 
\end{table*}

\begin{table*}[t]
\vspace{6mm}
\centering
\vspace{3mm}
\caption{Performance Comparison of the proposed method with common conventional method as well as other deep neural networks based Hazmat detection techniques}
\label{tab:comparison}
\begin{tabular}{|l|c|c|c|c|c|c|} \hline
						  \multirow{2}{*}{}& \multirow{2}{*}{\textbf{SIFT}\cite{SIFT2004} \%} & \multirow{2}{*}{\textbf{SURF}\cite{SURF2008} \%} & \multirow{2}{*}{\textbf{YOLOv2} \cite{redmon2016yolo9000} \%} & \multirow{2}{*}{\textbf{YOLOv3} \cite{yolov3} \% }& \multirow{2}{*}{\textbf{Proposed Method} \% } \\
						  & & & & & \\ \hline
Average Recall Rate       & 26.42 & 7.72 & 99.00 & 99.00 & 99.00 \\ 
Average Precision Rate    & 64.05 & 31.26 & 92.00 & 95.00 & 94.00 \\ 
Average IOU	Rate			  & 75.98 & 64.71 & 80.86 & 85.75 & 81.83 \\ 
mAP Rate				  & 64.05 & 31.26 & 99.70 & 99.37 & 99.03 \\ 
F1-Score      		      & 33.80 & 10.64 & 95.00 & 97.00 & 96.00 \\ \hline
Adaptive Bounding Box      & $\times$ & $\times$ & $\times$ & $\times$ & \textbf{\checkmark} \\ 
Multiple Object Detection & $\times$ & $\times$ & \checkmark & \checkmark & \textbf{\checkmark} \\ 
Segmentation              & $\times$ & $\times$ & $\times$  &  $\times$  &\textbf{\checkmark}  \\ \hline
Overall Speed							& $_{+}$ & $_{+}$ & $_{+++}$ &  $_{++}$ & $_{++++}$ \vspace{1mm}\\ \hline
\end{tabular} 
\end{table*}

\section{Conclusions}\label{conclusion}
In this paper we presented a robust system that can localise, classify, and segment Hazmat signs in the hazardous rescue fields.
The proposed methodology enabled us to confidently detect the presence of hazardous materials signs, regardless of the particular lighting situation, over a wide range of distances and under varying degrees of rotation. The trained model is also able to detect occluded, overlapped, and partially visible signs. The experimental results showed that the DeepHAZMAT model is more accurate and faster than many other recent and state-or-the-art research works such as \cite{hazmat-surf-cnn-2019}, \cite{CNN-uav}, and \cite{soren-cnn-2019}. The developed DNN-based system was fast enough to be implemented in Mobile robots, using a single Intel NUC Corei7 embedded system for robust and real-time hazard label detection, recognition, identification, localisation, and segmentation, thanks to skipping redundant input frames as well as adaptation of the YOLOv3-tiny for our real-time robotics application. As possible future work we suggest developing an optical character recognition method for text recognition inside the hazmat signs and to detect whole hazmat signs without selecting the background areas. In the interest of reproducible science and research, we have publicly released our unique dataset as well as the implemented code in the GitHub for the benefit of other researchers in the field. To the best of our knowledge we are the first that publish such a large and comprehensive dataset of HAZMAT signs with the ground truth annotations, to the global robotics community.

\medskip

\bibliographystyle{unsrt}
\bibliography{citation}
\end{document}